\documentclass[fleqn,10pt]{wlscirep}
\usepackage[utf8]{inputenc}
\usepackage[T1]{fontenc}
\title{Privacy-Preserving Federated Learning via Differential Privacy and Homomorphic Encryption for Cardiovascular Disease Risk Modeling}

\author[1,*]{Gaurang Sharma}
\author[1]{Juha Pajula}
\author[4]{Aada Illikainen}
\author[4]{Markus Rautell}
\author[2]{Noora Lipsonen}
\author[3]{Petri Alhainen}
\author[4]{Mika Hilvo}
\affil[1]{VTT Technical Research Centre of Finland Ltd, 33720, Tampere, Finland}
\affil[2]{Solita Oy, 00100 Helsinki, Finland}
\affil[3]{Mediconsult Oy, 00180 Helsinki, Finland}
\affil[4]{VTT Technical Research Centre of Finland Ltd, 02150, Espoo, Finland}

\affil[*]{gaurang.sharma@vtt.fi}


\usepackage{cite}
\usepackage{amsmath,amssymb,amsfonts}
\usepackage{graphicx}
\usepackage{textcomp}
\usepackage{xcolor}
\usepackage{makecell}
\usepackage{caption}
\usepackage{subcaption}
\usepackage{hyperref}
\usepackage{multirow}
\usepackage{amsmath} 
\usepackage{amsmath}
\usepackage{algorithm}
\usepackage{algpseudocode}
\usepackage{comment}
\usepackage{tabularx}
\usepackage{booktabs}

\usepackage{xcolor}

\begin{abstract}
Protecting sensitive health data while enabling collaborative analysis is a central challenge in healthcare. Traditional machine learning (ML) requires institutions to pool anonymized patient records, centralizing analytical development and privacy risks at a single site. Privacy-enhancing technologies (PETs), including Differential Privacy (DP) and Homomorphic Encryption (HE), can mitigate these risks. However, they are mainly studied in conventional data-sharing settings and often introduce trade-offs, including reduced model utility, higher computational cost, and increased implementation complexity. Federated Learning (FL) reduces data centralization by enabling institutions to train models locally and share only model updates. Nevertheless, FL does not eliminate privacy risks, as shared parameters or gradients may still reveal sensitive information. Integrating DP or HE into FL can strengthen privacy guarantees, yet their comparative performance and deployment implications in real-world healthcare settings remain unclear.

We systematically evaluated DP and HE integration in FL under real-world conditions, comparing them with standard FL and centralized ML (cML) to quantify privacy-utility trade-offs in multi-institutional settings. Using nationwide Swedish healthcare data, we evaluated cardiovascular disease risk prediction using logistic regression (LR) and neural network (NN) learners. FL with HE achieved performance comparable to cML but introduced measurable cryptographic overhead, particularly in the NN implementation. FL with DP incurred lower computational cost; however, LR was more sensitive to calibrated noise than the NN, resulting in greater performance degradation. Our findings provide practical guidance for deploying privacy-preserving FL in fragmented healthcare systems.

\end{abstract}
\begin{document}

\flushbottom
\maketitle

\thispagestyle{empty}

\section*{Introduction}


Exposure of sensitive personal information creates serious privacy risks. Privacy-enhancing technologies (PETs)\cite{khan2026privacy} help mitigate these risks by limiting data exposure while enabling data analysis. However, privacy is never absolute. Protecting privacy requires continuous adaptation to evolving threats and often involves trade-offs between privacy, utility, and computational performance. These trade-offs are especially consequential in healthcare, where data is highly sensitive and often distributed across multiple institutions and systems. Traditional healthcare data analysis workflows often rely on transferring data to a centralized platform, which increases privacy risks and complicates regulatory compliance when data crosses organizational or jurisdictional boundaries. As a result, this approach centralizes not only the analysis but also the privacy risks.

Once data are transferred from their source, the risks of leakage, unauthorized access, misuse, and re-identification increase substantially. To mitigate these risks, PETs such as Differential Privacy (DP) and Homomorphic Encryption (HE) have been developed and added to traditional workflows~\cite{khan2026privacy, dwork2006calibrating, acar2018survey}. DP limits inference about individuals by bounding the influence of any individual record on the output, often through calibrated noise, while HE enables computations directly on encrypted data, producing encrypted outputs that only authorized parties can decrypt. However, applying these techniques within centralized workflows does not eliminate the risks associated with pooling sensitive data. Federated Learning (FL)\cite{mcmahan2017communication} offers an alternative by allowing analysis or training of ML models in a distributed setting. In FL, each data source performs local computations and shares only the trained model parameters to support global model training. Because data never leave their source, FL reduces the risks associated with centralized data analysis. However, parameter sharing introduces new vulnerabilities \cite{guo2026combining, MA2026131503}, since adversaries can potentially, e.g., reverse-engineer shared model updates to infer original sensitive information. Thus, combinations of different PETs, e.g., FL with DP or HE, are gaining attention to maximize privacy. 

When DP is integrated into FL (FL\_DP), each data source clips the trained model updates (e.g., gradients) and adds calibrated noise before sharing them. When HE is integrated (FL\_HE), a data source encrypts its model updates so that the server can perform analyses without ever accessing the plaintext values, the server aggregates the encrypted trained parameters received from all sources and shares them with each source, and eventually the source decrypts the aggregated updates for continued training. Thus, incorporating DP or HE into FL does not allow the sharing of raw trained parameters; it strengthens privacy guarantees and makes FL particularly attractive for healthcare applications. However, these approaches introduce trade-offs: FL\_HE incurs both computational and communication overhead, while FL\_DP may reduce model performance in exchange for stronger privacy.

Although prior studies have investigated DP~\cite{cheng2022dpnas, pannekoek2021investigating, negoescu2023epsilon, wei2023dpmlbench} and HE~\cite{matias2023exploring, minelli2018fully}, and even have combined them with FL~\cite{app12020734, saifullah2024towards, tang2018homomorphic, 10964604, 10320195}, to our knowledge no previous work has systematically compared FL\_DP and FL\_HE in real-world healthcare deployment settings. Moreover, most studies have focused exclusively on NN, leaving traditional ML methods underexplored. We address these gaps by evaluating both approaches in a cardiovascular disease (CVD) prediction task. Using Swedish healthcare data, we aim to predict whether an individual will experience a major CVD event within four years after analyzing their medical history, medications, and demographic variables. We simulated, deployed, and analyzed four learning paradigms: (1) centralized machine learning (cML) on pooled data, (2) standard Federated Averaging (FedAvg), (3) FedAvg with DP (FedAvg\_DP), and (4) FedAvg with HE (FedAvg\_HE). We used cML as a performance benchmark to assess whether FL achieves comparable predictive accuracy. We then compared federated approaches to analyze convergence under DP and HE, and to quantify privacy–utility trade-offs in a realistic healthcare deployment scenario.

\section*{Literature review}
\subsection*{Federated learning}
\label{FL}
FL\cite{mcmahan2017communication} is a distributed ML approach enabling local model development through collaboration without sharing sensitive data. First, the server initializes and shares the global model with data sources, which then perform local training (epochs). The sources then share their model parameters with a central server, which aggregates and updates the global model. The updated global parameters are then sent back to the source for further local training. This cycle repeats until convergence.

FL can be categorized in various ways \cite{manzoor2024survey}. With respect to data partitioning, there are three main categories: horizontal FL, vertical FL, and federated transfer learning. 
Another way to categorize FL schemes is based on system architecture, dividing them into centralized FL (client-server) and decentralized FL (peer-to-peer) \cite{zhang2021survey}. In the centralized FL setup, all clients sends the model parameters to the server and the server responds back by sharing the aggregated parameters. The decentralized setup is the opposite, where clients directly communicate and share updated model parameters with each other. This method improves both resilience and privacy by eliminating a single point of failure and avoiding a central data repository. However, it necessitates more sophisticated coordination protocols to handle peer-to-peer communication.
The third categorization approach is based on operational strategies, i.e., on cross-device or cross-silo\cite{jiang2021flashe}. 
The first setting, i.e., cross-device FL, typically involves numerous less powerful devices, such as mobile phones and IoT devices, that collect personalized data. The latter, i.e., cross-silo FL, typically refers to scenarios in which participants are organizational entities and have proportionately large population datasets.

One of the main advantages of FL is that it does not require sharing sensitive local data, but like all other ML methods, FL has its challenges that can be broadly divided into two types: training bottlenecks and privacy and security concerns. The training bottlenecks include communication overhead as well as system and data heterogeneity \cite{zhang2021survey}. The recognized privacy and security concerns can be divided into three main categories \cite{manzoor2024survey}: data, model and privacy attacks. Data attacks, such as data poisoning \cite{liu2022threats}, label flipping \cite{lv2022awfc} and backdoor attacks \cite{mammen2021federated,andreina2021baffle}, target the training data at source, to negatively impact the learning process.
In contrast, during a model attack, malicious sources alter their local model updates to corrupt the global model by introducing harmful changes \cite{zhou2021deep}, often by modifying parameters to produce incorrect or biased outputs. Privacy attacks exploit the fact that, despite the privacy-preserving nature of FL, the shared local model parameters can potentially leak sensitive information about a source’s local data, which can then motivate malicious actions.

As a result, it is preferable to combine the FL setup with suitable PETs such as HE, DP. Utilizing PETs introduces new challenges in the model training process, such as additional time and payload overheads, and a decrease in the accuracy of global model aggregation.
While there are existing solutions and countermeasures for these challenges, they remain an active area of research for future improvements and innovations.

\subsection*{Differential privacy}
\label{DP}
Differential privacy (DP) \cite{dwork2006calibrating} provides a formal framework for quantifying and safeguarding privacy in data analysis and ML. It acts as a countermeasure against malicious actors, for example those attempting to recover information about the original data set from a trained model \cite{ren2022grnn} or inferring whether a particular individual was included in the training data set \cite{shokri2017membership}. 
DP guarantees that the output of an algorithm is nearly indistinguishable whether or not any single individual's record is included in the dataset~\cite{dwork2006calibrating}. This ensures that the contribution of any individual record has a bounded and measurable effect on the released output, thereby limiting the information that can be inferred about any particular individual.

DP methods rely on the introduction of calibrated noise that may be applied at different stages of model development: before, during or after training \cite{el2022differential}. Equivalently, noise can be added to the original data, intermediate statistical queries, or released model outputs. DP mechanisms are commonly categorized into two models. In the central DP model, data sources or users place trust in a central database operator that performs noise addition before releasing results. In contrast, in the local DP model, the data source applies the noise locally before submitting any information to a centralized operator. 

The primary challenge in the practical deployment of DP mechanisms is the trade-off between privacy and utility. In ML applications, this utility is typically reflected in empirical model performance metrics, such as predictive accuracy or convergence behavior. In practice, stronger privacy protection generally requires increased noise injection, which can decrease model accuracy. Addressing this challenge requires careful calibration of privacy parameters to balance accuracy and privacy preservation. For further discussion of differential privacy in ML and FL, we refer the reader to \cite{adnan2022federated, dong2022gaussian, shen2022distributed, el2022differential}.

\subsection*{Homomorphic encryption}
\label{HE}

HE enables arithmetic operations to be performed directly on encrypted data, allowing third parties to carry out computations without accessing the plaintext values. In algebraic terms, a homomorphism is a structure-preserving mapping between two sets. HE schemes are commonly categorized based on the class of computations they support \cite{acar2018survey}. Partially homomorphic encryption (PHE) schemes allow evaluation of circuits containing a single type of operation, such as addition or multiplication. Somewhat homomorphic encryption (SWHE) schemes support both addition and multiplication, but only for a limited number of operations. Leveled fully homomorphic encryption (LFHE) schemes enable the evaluation of arbitrary circuits up to a predefined depth $L$ \cite{brakerski2014leveled}. Fully homomorphic encryption (FHE) represents the most general class, allowing the evaluation of arbitrary-depth circuits composed of multiple gate types. FHE schemes rely on bootstrapping, a technique introduced by Gentry~\cite{gentry2009fully}, to control noise growth in ciphertexts (encrypted data) and thereby support an unbounded number of evaluations on plaintexts (unencrypted data).

While HE provides strong confidentiality guarantees by ensuring that data remain encrypted throughout computation, it introduces substantial computational and communication overhead. Operations on ciphertexts are significantly slower than their plaintext counterparts, and the algebraic structure required to support homomorphic evaluation results in significant ciphertext expansion. These factors impact both runtime and communication costs, making HE most suitable for scenarios where strong security guarantees outweigh efficiency considerations.

Most widely used HE schemes operate on integer-valued inputs, but the CKKS (Cheon-Kim-Kim-Song) scheme is an important exception. Introduced by \cite{cheon2017homomorphic}, CKKS supports encrypting real and complex numbers and allows for approximate arithmetic over encrypted data. It was introduced as a leveled HE scheme, in which the available modulus decreases with each rescaling, thereby limiting the achievable multiplicative depth. Consequently, a bootstrappable variant \cite{cheon2018bootstrapping} was later proposed, extending CKKS to a fully homomorphic encryption (FHE) scheme. Subsequent work has introduced a range of optimizations to improve bootstrapping efficiency and numerical stability \cite{bossuat2022bootstrapping, chen2019improved, cheon2018faster, cheon2018full}. Collectively, these advances demonstrate that CKKS is not only of theoretical interest but also practical for real-world applications involving approximate arithmetic.

\section*{Methods}

We conducted a retrospective study using nationwide Swedish healthcare data to evaluate privacy-preserving ML approaches for CVD risk prediction. The workflow progressed from controlled simulation to real-world deployment. We first developed an in-house Secure-Health (SeH) platform to host the dataset and establish the baseline centralized ML pipeline. The data were then partitioned to simulate distributed data sources, enabling the development and simulation of an FL infrastructure within the platform. After validating the system in the simulated environment, the federated nodes were deployed to remote clients, with the server node hosted on the SeH platform.

\subsection*{Data}
The dataset was obtained from the National Board of Health and Welfare (Socialstyrelsen), a governmental agency under the Swedish Ministry of Health and Social Affairs responsible for maintaining national health registries. The dataset integrates records from three comprehensive nationwide registries: the National Patient Register, the National Prescribed Drug Register and the Register for Interventions in Municipal Health Care.
The dataset includes anonymized patient information from all regions of Sweden covering the period from January 2009 to November 2024.

We defined CVD diagnosis as composite endpoint comprising ICD‑10 codes I21 (myocardial infarction), I22 (subsequent myocardial infartion) and I50 (heart failure). We used each individual’s most recent recorded diagnosis to predict the risk of a CVD event within four years after the observation period cut-off date (31 December 2020). The raw dataset included 4,327,532 individuals and 2,536,803 diagnosis records and among these, 1,598,372 individuals had at least one recorded CVD diagnosis. We restricted the cohort to patients with complete pre‑cutoff medical histories and no CVD diagnoses before the end of 2020 along with the predicted the risk of incident CVD events after the cut-off date. This process yielded 42,207 patients labeled as a “CVD case” and 618,394 patients without CVD before or after the cutoff, labeled as a “Non‑CVD case,” resulting in a final cohort of 660,427 participants with 10 predictor variables. The predictors included diabetes (E10–E14), disorders of lipoprotein metabolism and other dyslipidemias (E78), demographic information (age and gender), and medication codes: ATC\_A10 for antidiabetics, ATC\_09 for RAAS agents (Renin-Angiotensin-Aldosterone system inhibitors) and ATC\_C10 for lipid-modifying drugs. More details on the dataset and its preprocessing are found in our recent publication \cite{sharmaSecurEhealth}.

\subsection*{Model implementations}
The study used two learning methodologies: cML and FL.
The cML methodology used the full dataset available on the SeH platform to train two learners: a logistic regression (LR) model and an artificial neural network (NN), each evaluated with 10-fold cross-validation. In each iteration, one fold served as the test set, while the remaining nine folds served as the training set.
The LR learner used the Scikit-learn ML library in Python~\cite{scikitlearn}. The liblinear solver optimized the regularized LR model with default settings\footnote{\url{https://scikit-learn.org/stable/modules/generated/sklearn.linear_model.LogisticRegression.html}}
. The LR learner comprised 11 trainable parameters: ten feature coefficients and one intercept term.
The NN learner used a custom PyTorch model. The architecture consisted of a lightweight feed-forward network that processed ten input features through a single hidden layer with five neurons, followed by ReLU activation and layer normalization to improve training stability. A single-neuron output layer produced the logit for binary classification. Xavier uniform initialization initialized all weights, and all biases were initialized to zero to support stable convergence. The NN learner comprised 66 trainable parameters, including weights, biases, and learnable layer-normalization parameters.

The second learning methodology employed horizontal FL in a cross-silo setting, using a centralized client-server architecture. To reflect a realistic FL scenario, the full dataset was partitioned across four Swedish counties: Stockholm, Uppsala, Södermanland, and Östergötland, according to population statistics on age distribution, gender, and total population in 2020 (see Table~\ref{tab:cvd_distribution} for data distribution).
The SeH server securely distributed the dataset to the corresponding clients~\cite{sharmaSecurEhealth}. Each client then locally partitioned its data into training (80\%) and validation (20\%) subsets.

To ensure comparability across methodologies, the same learners used in the cML setting were employed as local FL learners. The FL methodology included three implementations: FedAvg, FedAvg\_DP, and FedAvg\_HE. In the standard FedAvg implementation, clients trained local models on plaintext data and transmitted model updates to the server for aggregation, without applying any additional privacy mechanisms.
FedAvg\_DP followed the same procedure as FedAvg but incorporated a differential privacy mechanism based on the SVTPrivacy algorithm. This approach perturbed local model updates with calibrated noise prior to transmission, thereby enhancing privacy protection.
In FedAvg\_HE, local model updates were encrypted at each client using the CKKS scheme before being sent to the server. Aggregation was performed directly on the encrypted updates, and the resulting global model update was returned in encrypted form. Clients then decrypted the aggregated update locally.
Model updates refer to the parameter differences (deltas) between model weights before and after local training at each client.

In all FL methodologies, each client trained its selected local learner, either LR or NN, using only its private training data. At each federated round, the local validation set evaluated both the received global model and the updated local model. Global model performance was evaluated before local training, whereas local model performance was evaluated after local training.
After all global training rounds, cross-site validation was performed by evaluating the global model on each site-specific validation set.


\begin{table}[!htbp]
\centering
\begin{tabular}{|l|r|r|r|}
\hline
\textbf{Client site} & \textbf{Non-CVD cases} & \textbf{CVD cases} & \textbf{Total} \\
\hline
Östergötland & 92{,}630 & 6{,}518 & 99{,}148 \\
Södermanland & 63{,}901 & 4{,}575 & 68{,}476 \\
Stockholm    & 391{,}954 & 26{,}046 & 418{,}000 \\
Uppsala      & 69{,}909 & 4{,}894 & 74{,}803 \\
\hline
\end{tabular}
\caption{Distribution of Non-CVD and CVD individuals across client sites.}
\label{tab:cvd_distribution}
\end{table}

\subsection*{Privacy-preserving mechanisms for FL}

Differential privacy was incorporated into the FL workflow by applying a Sparse Vector Technique (SVT)–based privacy mechanism into the client-side communication pipeline before transmission of model updates to the server (local DP model), yielding the FedAvg\_DP method.   
The SVTPrivacy mechanism applied a norm-based query to each update, selectively perturbing and releasing its components using the Sparse Vector Technique. It released only components that passed a noisy threshold, added Laplace noise, and limited the number released to control privacy loss.
In the FedAvg\_DP implementation, clients enforced DP by applying Laplace noise to their model updates using SVT (details found on NVFLARE documentation \footnote{\url{https://nvflare.readthedocs.io/en/2.4.1/apidocs/nvflare.app_common.filters.svt_privacy.html}}). Specifically, the noise scale parameter and the privatized threshold were
computed jointly as
\begin{align}
    \lambda_{\rho} &= \frac{2\gamma}{\varepsilon}, \label{eq:noise_scale}\\[4pt]
    \text{threshold} &= \tau + \operatorname{Laplace}(0,\,\lambda_{\rho}),
    \label{eq:threshold}
\end{align}
where $\gamma$ denotes the clipping bound, $\varepsilon$ is the privacy budget, and $\tau$ defines the baseline selection threshold. The noise scale
$\lambda_{\rho}$ in Equation~\eqref{eq:noise_scale} thus governs the magnitude of the
Laplace perturbation applied to the threshold in Equation~\eqref{eq:threshold}.

At each communication round, clients flattened their local model updates, normalized them by the number of local optimization steps, and clipped them to the bound~$\gamma$. A noisy thresholding mechanism was then applied to retain only those parameters whose magnitudes exceeded the privatized threshold. The privacy budget parameter~$\varepsilon$ controlled the overall noise injected during parameter selection, while the threshold offset~$\tau$ determined the baseline selection criterion prior to noise injection. The parameter \textit{fraction} specified the proportion of update elements retained and transmitted. Subsequently, Laplace noise with variance controlled by \textit{noise\_var} was added to the selected parameters. The privatized updates were reshaped back to their original tensor structure and rescaled before transmission to the server. The remaining configuration parameters, \textit{data\_kinds} and \textit{replace}, were kept at their default values (\texttt{None} and \texttt{True}, respectively), as defined in the NVFLARE implementation.\footnote{\url{https://nvflare.readthedocs.io/en/2.4.1/apidocs/nvflare.app_common.filters.svt_privacy.html}}

Homomorphic encryption was implemented using the CKKS scheme~\cite{cheon2017homomorphic} to safeguard model updates, yielding the FedAvg\_HE algorithm. The CKKS scheme was implemented with a polynomial degree of $8192$, coefficient modulus sizes of $[60, 40, 40]$ and scaling factor of $2^{40}$. Prior to transmission, each client encrypted its update using the CKKS public key, producing an encrypted update \(\mathsf{Enc}(\Delta \boldsymbol{\theta}_i).\)
The encrypted client updates were transmitted to the server, which computed the encrypted aggregate.
The aggregation was performed by accumulating the client model
parameters and normalizing by the sum of client weights,
\begin{equation}
    \Delta\theta_{\text{global}}
    = \frac{\displaystyle\sum_{i=1}^{N} \Delta\theta_{i}}
           {\displaystyle\sum_{i=1}^{N} w_i},
    \label{eq:fedavg_agg}
\end{equation}
where $w_i$ denotes the weight assigned to client~$i$. This formulation is
equivalent to standard FedAvg when the client updates are scaled consistently
with $w_i$ prior to aggregation. Under the assumption of equal client
contributions, $w_i = 1$ for all~$i$, so that $\sum_{i=1}^{N} w_i = N$ and
the aggregation reduces to a simple average. All model parameters remained
encrypted throughout the aggregation process.

Since direct division was not supported in the encrypted domain \cite{benaissa2021tenseal}, NVFLARE avoided ciphertext division by postponing the application of aggregation weights until the final normalization step, thereby reducing the number of costly ciphertext–plaintext multiplications. The server first computed an unweighted encrypted sum,

\[
    \sum_{i=1}^N \mathsf{Enc}(\Delta\theta_i),
\]
and then multiplied this sum by a plaintext reciprocal to obtain the aggregated
global model,

\[
    \mathsf{Enc}(\Delta\theta_{\text{global}})
    =
    \Bigg(\sum_{i=1}^N \mathsf{Enc}(\Delta\theta_i)\Bigg)
    \cdot
    \frac{1}{\displaystyle\sum_{i=1}^{N} w_i}.
\]
This aggregation strategy avoided ciphertext division entirely and required only ciphertext additions during accumulation and a single ciphertext–plaintext multiplication per model parameter during normalization \cite{nvflare-weighted-aggregation-helper}.
After aggregation, the encrypted global update was returned to the clients (or a designated decryptor), who decrypted it with the private key to recover $\Delta\theta_{\text{global}}$ in plaintext.

\subsection*{SeH platform}

The SeH platform was developed as a unified infrastructure that enables researchers to perform data analysis, build ML models as well as simulate and deploy FL infrastructure in centralized or distributed environments. We deployed the SeH Platform on a virtual server configured as a Docker host with direct internet connectivity. The virtual server had 32GB of RAM, 32 CPU threads (Intel(R) Xeon(R) Platinum 8380 CPU @ 2.30GHz), and 1 TB of storage, and was running on a large KVM host. All applications and tools were executed inside Docker containers\footnote{Docker containers \url{https://www.docker.com/resources/what-container/}}, and managed directly by administrators or users using predefined Docker images via the ShinyProxy\footnote{ShinyProxy \url{https://shinyproxy.io/}} application. ShinyProxy, with fully containerized deployment, provided a web portal interface that allowed users to launch applications based on group memberships. In ShinyProxy, each started application runs on an isolated container with a web interface embedded in an Iframe within the Shinyproxy web interface.

Platform communication was secured by a containerized NGINX HTTPS proxy that routed traffic to the ShinyProxy container while enforcing HTTPS and internal network restrictions. Authentication of the ShinyProxy web interface was handled via Microsoft Entra ID\footnote{Entra ID: \url{https://www.microsoft.com/en-us/security/business/identity-access/microsoft-entra-id}} deployed at a separate MS Azure cloud tenant, enabling role-based access control. Users were segmented into three groups, ranging from administrators to basic users with limited access. Data within containers remained ephemeral, with applications configured to terminate after periods of inactivity. Persistent storage was mounted on a per-user and per-application basis to ensure session continuity, while shared storage enabled users to exchange results. Additional containerized services included SFTP, PostgreSQL, and NVFLARE components for data management and external connectivity. All containers were running inside a closed Docker NAT network, except the NGINX HTTPS proxy, the SFTP server, and the NVFLARE server, which had internet connectivity via port forwarding and a dedicated firewall rule.

The SeH platform was configured to host the FL server along with two client nodes containing the Södermanland and Östergötland datasets. These nodes were deployed in isolated containers that shared computational resources with the FL server and the administrative process containers, thereby enabling real-world FL execution. The remaining client nodes were hosted independently by Mediconsult and Solita in geographically distinct regions, each with heterogeneous hardware configurations. Solita deployed its node using the Uppsala dataset on a Google Cloud \texttt{e2-medium} instance (2~vCPUs, 4~GB RAM, shared-core) running on an x86\_64 (amd64) architecture. Mediconsult deployed its node using the Stockholm dataset on a Microsoft Azure \texttt{Standard B2als v2} virtual machine equipped with 2~vCPUs (AMD EPYC\texttrademark~7763, x64 architecture) and 4~GB of RAM. Together, these independently deployed nodes validated distributed model training across organizational boundaries over the public internet.

\subsection*{Simulation and real-world FL}

The FL implementations were first evaluated in a controlled simulation environment before deployment in a real-world infrastructure. Within the dedicated project environment on the SeH platform, the simulation emulated a client-server architecture by assigning distinct IP addresses to each client and the central server to ensure network isolation. During simulation, each client loaded its local dataset and participated in collaborative model training. Iterative hyperparameter tuning was used to optimize the NN and LR models. Each client performed local training for 20 epochs per round across 250 global communication rounds. The final configuration used a learning rate of 0.01 and a batch size of 20,000 for all clients except Stockholm, which used a batch size of 100,000.

After completing the baseline FedAvg simulation, DP and HE mechanisms were incorporated into the training pipeline. Privacy and encryption parameters were then calibrated to ensure robustness, computational feasibility, and suitability for real-world deployment. Following successful simulation, the FL system was deployed in a production environment. Provisioning kits were generated from a preconfigured project.yml specification containing digital certificates, configuration files, and communication endpoints. Client datasets and startup provisioning kits were securely distributed to federated client sites operating on independent hardware across diverse IT environments. The server startup kit was deployed on the designated SeH platform server and configured to accept connections only from whitelisted Internet IP addresses.

During provisioning, each client initiated a worker instance and submitted a join request with its certificate and metadata for authentication. The FL server verified client identities through mutual Transport Layer Security (mTLS) and issued authentication tokens after successful validation, enabling secure participation in the federation. After provisioning, the administrator submitted an FL job package through the administrative interface. The FL server then deployed the application to authenticated clients, initialized the global model using the configured model persistor, and distributed the initial model parameters along with workflow instructions, task configuration, and hyperparameters. The federated environment then executed the finalized model configurations, optimized during simulation, enabling coordinated local training and model aggregation across participating clients.

\section*{Results}

\subsection*{Computational performance across implementations and learners}

\begin{figure}
    \centering
    \includegraphics[width=1\linewidth]{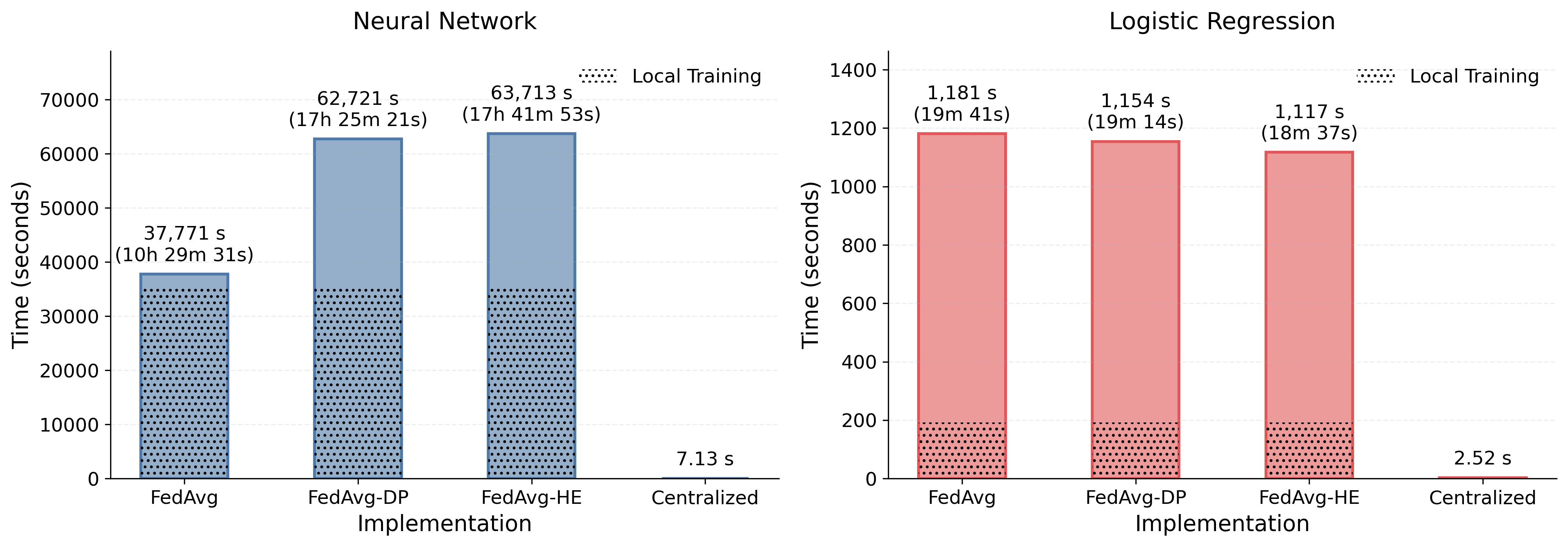}
    \caption{Total model training time encompassing 250 global epochs combined with 20 local epochs, with all models utilizing identical hyperparameters and architectures.}
    \label{fig:computationalTime}
\end{figure}

Figure \ref{fig:computationalTime} demonstrates the end-to-end runtime of each method. 
Among the federated implementations, FedAvg achieved the lowest runtime using the NN learner and a slightly higher runtime than other FL implementations using the LR learner. With the NN learner, FedAvg completed 250 global rounds with 20 local epochs per round in 37,771~s (10~h~29~m~31~s), whereas FedAvg\_DP and FedAvg\_HE required 62,721~s (17~h~25~m~21~s) and 63,713~s (17~h~41~m~53~s), respectively. FedAvg required approximately 7 hours less computational and communication time compared to other federated privacy-preserving variants. FedAvg\_HE exceeded FedAvg\_DP by 992~s (16~m~32~s), reflecting the additional overhead of HE.
With the LR learner, runtimes were almost similar: 1,181~s (FedAvg), 1,154~s (FedAvg\_DP), and 1,117~s (FedAvg\_HE). The differences are negligible relative to total runtime and likely attributable to system variability rather than algorithmic overhead.
The decomposition shows that local training accounts for 192.5~s with LR and 35,527.5~s (9.87~h) with NN, indicating that computation dominates the NN runtime, whereas communication and coordination constitute a larger relative share of the LR runtime.

In detail, with LR, the mean local training time per round was 0.77~s at the Stockholm and 0.15~s at the remaining client sites, with the server observing client update arrivals after 4.61~s and 1.64~s, respectively. With the NN, the local training required on average 142.11~s at Stockholm and 34.82~s at the other client sites, while server-side update arrival times were 246.3~s and 54.43~s, respectively. 
Across all federated implementations, server-side processing remained negligible: handling each client update required less than 0.12~s, validation and registration required approximately 0.21~s per client, and aggregation commenced about 0.20~s after the final update was received. Because aggregation was triggered only after all client updates arrived, faster clients experienced idle time while waiting for the slowest participant.

The centralized implementations were substantially faster, completing in 2.52~s (LR) and 7.13~s (NN), as they avoid communication, synchronization, and privacy-related overhead.

\subsection*{Privacy and performance tradeoff in differential privacy}

\begin{figure}[htbp]
    \centering
    \includegraphics[width=0.9\linewidth]{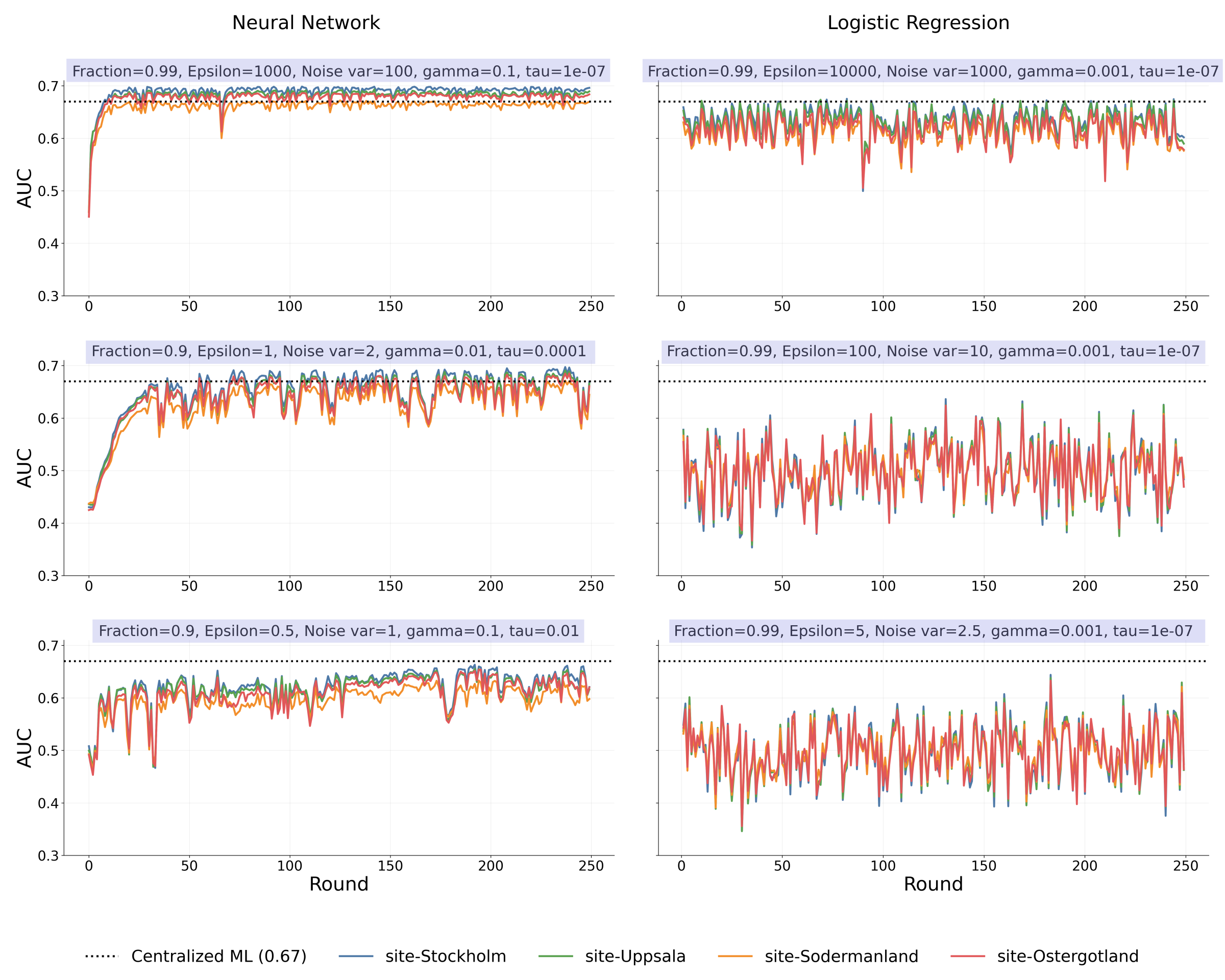}
    \caption{Effect of privacy configurations on the global model's performance evaluated on each site’s local validation set for different FL learners. With a higher privacy budget and lower noise level, the NN-based model (66 trainable parameters) closely matched the standard FedAvg implementation (see Fig.~\ref{fig:method_level_comparison}), whereas the LR-based model (11 trainable parameters) remained more sensitive to noise and showed performance degradation even at lower noise levels. Relaxing privacy constraints improved model utility at the cost of reduced privacy protection, highlighting the privacy–utility trade-off.}
    \label{fig:dp_fineTuning}
\end{figure}

We evaluated multiple configurations for each FL learner (NN and LR) to assess the trade-off between privacy and model performance. Figure~\ref{fig:dp_fineTuning} visualizes some configurations illustrating the effect of noise settings on global parameters. Relaxing DP constraints, corresponding to a higher privacy budget and lower noise level, led to smoother convergence and minimal performance degradation. In contrast, stricter privacy constraints introduced higher-calibrated noise, resulting in less stable training and lower predictive performance. Moreover, lighter local FL learners, i.e., LR with 11 trainable parameters, were more sensitive to DP noise than the NN-based learner with 66 trainable parameters, highlighting that privacy configurations are not directly transferable across learners and must be tuned according to model size and architecture. 

Considering the observed trade-off between privacy protection, model utility, and tuning complexity, we selected $\mathrm{Fraction}=0.9$, $\epsilon=1$, $\mathrm{NoiseVar}=2$, $\gamma=0.01$, and $\tau=10^{-4}$ for the NN-based FedAvg\_DP implementation. For the LR-based FedAvg\_DP implementation, we selected $\mathrm{Fraction}=0.99$, $\epsilon=10^{4}$, $\mathrm{NoiseVar}=1000$, $\gamma=0.001$, and $\tau=10^{-7}$.

\subsection*{Privacy and efficiency trade-off in homomorphic encryption}

Unlike FedAvg\_DP, FedAvg\_HE does not inject noise into model parameters and therefore does not directly impact predictive performance. Minor numerical deviations arise only from the fixed-precision arithmetic of approximate homomorphic schemes such as CKKS, rather than from privacy-induced perturbations. Consequently, the primary trade‑off of FedAvg\_HE lies in increased computational and communication costs, while model utility is largely preserved apart from effects due to numerical approximation and algorithmic adaptations required for homomorphic evaluation.

Computational performance analysis confirmed that FedAvg\_HE introduces measurable cryptographic overhead compared to standard FedAvg and incurs additional costs to encrypt, transmit, and aggregate model updates relative to FedAvg\_DP. This overhead reflects both the time required for encryption and decryption and the size of the encrypted model updates. For LR, encryption and decryption each required approximately 0.005~s, and the encrypted model updates measured 1.8~KB. In the NN setting, encrypted updates increased to approximately 5.4~MB, with encryption and decryption requiring 0.013~s and 0.007~s, respectively.

\begin{figure}
    \centering
    \includegraphics[width=0.9\linewidth]{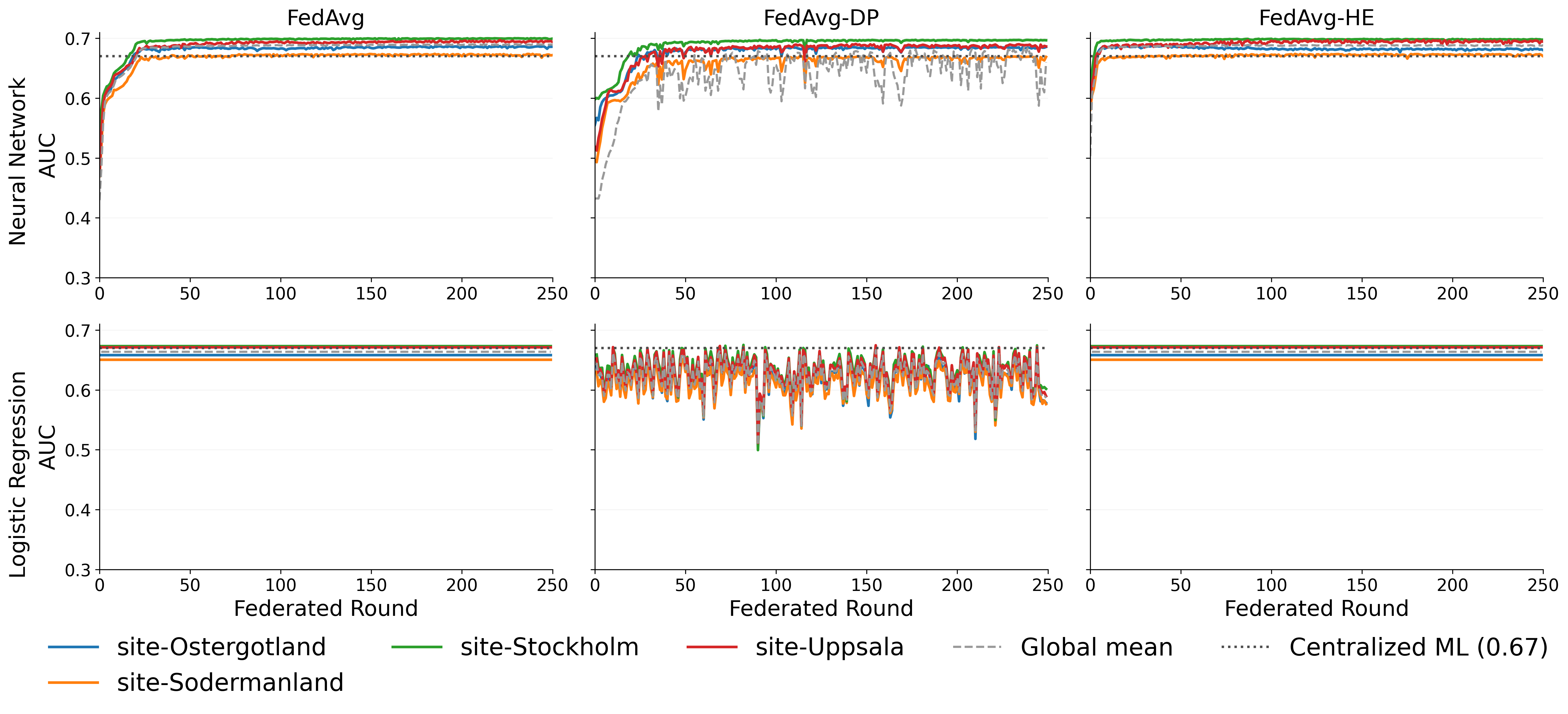}
    \caption{FedAvg\_HE and standard FedAvg converged faster and achieved more stable scores, compared to FedAvg\_DP. In FedAvg\_DP, the model's performance varies noticeably, attributed to calibrated noise. The curves demonstrate each site’s local model performance on its corresponding local validation set across federated rounds. Global mean represents the average performance of the global model across all sites' local validation sets.}
    \label{fig:method_level_comparison}
\end{figure}

\subsection*{Federated methods achieved comparable performance to that of cML}

cML achieved an AUC of 0.67 in the 10-fold cross-validation. Federated implementations demonstrated performance comparable to cML, except for the LR learner in
FedAvg\_DP, which showed performance degradation due to greater sensitivity to DP noise. Figure~\ref{fig:method_level_comparison} illustrates the performance of local and global models on local validation sets. With the NN learner, local models converged within 20 global rounds in FedAvg and FedAvg\_HE, whereas in FedAvg\_DP, convergence occurred after around 50 rounds. In contrast, the LR learner yielded stable performance across all rounds in FedAvg and FedAvg\_HE implementations. This consistent performance was attributed to the larger batch size. However, variation and performance degradation were observed in FedAvg\_DP due to the model's sensitivity. 

\begin{figure}
    \centering
    \includegraphics[width=0.9\linewidth]{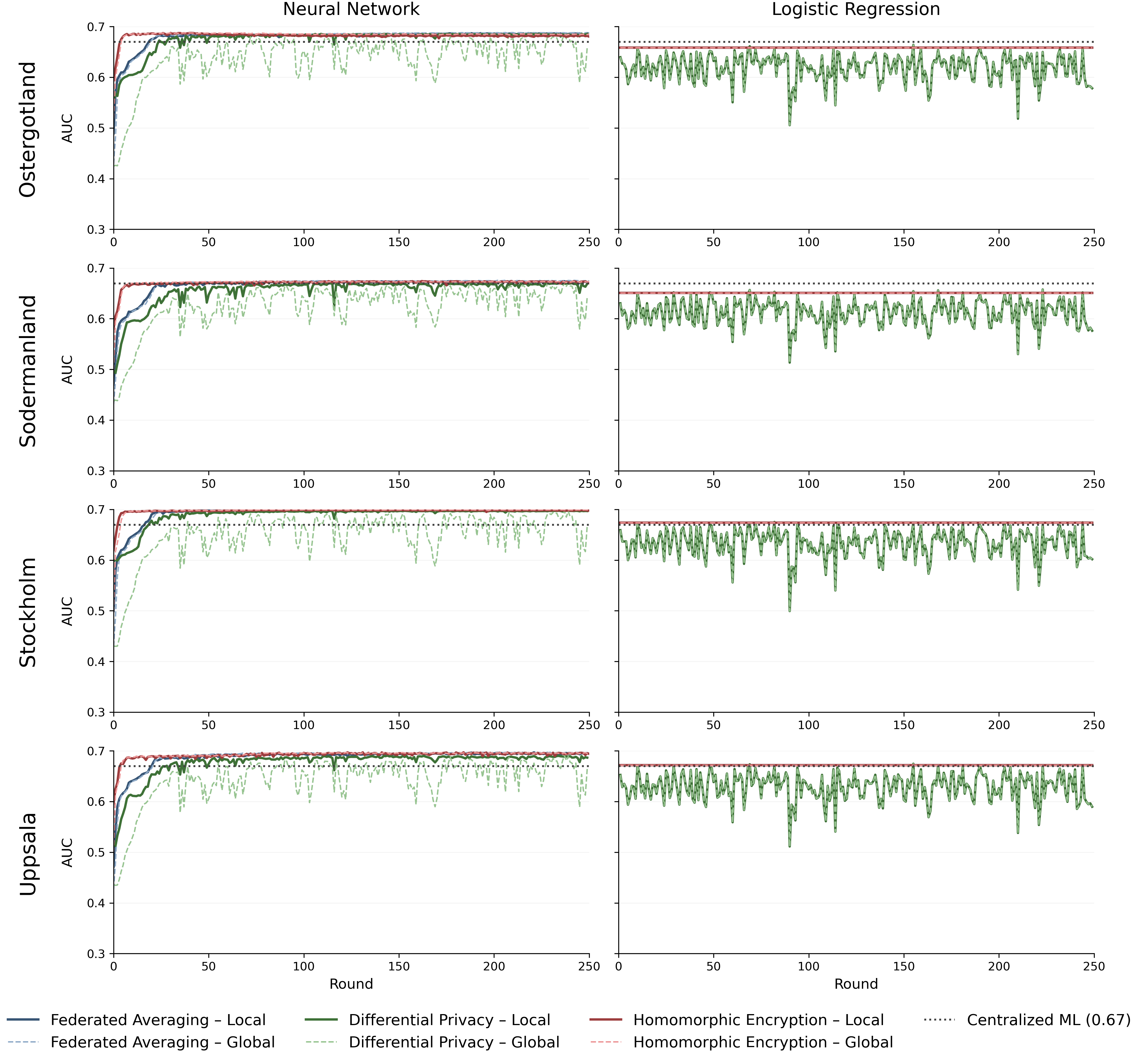}
    \caption{Site-specific performance on local validation sets across FL implementations and local learners. FedAvg\_HE showed the fastest convergence, followed by standard FedAvg and FedAvg\_DP. For FedAvg and FedAvg\_HE, global and local model performance were similar. In contrast, FedAvg\_DP showed a noticeable difference between global and local model performance within the NN-based implementation. The local model was more stable, as performance was recorded after local training.}
    \label{fig:Site_specific_scores}
\end{figure}

Figure~\ref{fig:Site_specific_scores} illustrates similar results from the site's perspective. Table~\ref{tab:global_model_privacy_methods} reports the mean and standard deviation of cross-validation scores of the global model across all local validation sets.

\begin{table}[ht]
\centering
\renewcommand{\arraystretch}{1.2}
\begin{tabularx}{\textwidth}{|l|p{3.6cm}|l|X|X|X|}
\hline
\textbf{Method} & \textbf{Implementation} & \textbf{Local Learner} & \textbf{AUC (mean $\pm$ std)} & \textbf{Sensitivity (mean $\pm$ std)} & \textbf{Specificity (mean $\pm$ std)} \\
\hline

\multirow{6}{*}{\textbf{FL}} 
& \multirow{2}{*}{\makecell[l]{Federated Averaging\\(FedAvg)}}
    & Neural Network & 0.680 $\pm$ 0.012 & 0.184 $\pm$ 0.034 & 0.897 $\pm$ 0.054 \\
&   & Logistic Regression & 0.663 $\pm$ 0.009 & 0.204 $\pm$ 0.045 & 0.892 $\pm$ 0.025 \\
\cline{2-6}

& \multirow{2}{*}{\makecell[l]{Differential Privacy\\(FedAvg\_DP)}}
    & Neural Network & 0.657 $\pm$ 0.018 & 0.079 $\pm$ 0.024 & 0.896 $\pm$ 0.058 \\
&   & Logistic Regression & 0.623 $\pm$ 0.028 & 0.018 $\pm$ 0.071 & 0.988 $\pm$ 0.045 \\
\cline{2-6}

& \multirow{2}{*}{\makecell[l]{Homomorphic Encryption\\(FedAvg\_HE)}}
    & Neural Network & 0.686 $\pm$ 0.011 & 0.190 $\pm$ 0.045 & 0.901 $\pm$ 0.006 \\
&   & Logistic Regression & 0.664 $\pm$ 0.010 & 0.205 $\pm$ 0.045 & 0.893 $\pm$ 0.026 \\
\hline

\multirow{2}{*}{\textbf{ML}} 
& \multirow{2}{*}{Centralized (cML)}
    & Neural Network & 0.672 $\pm$ 0.030 & 0.166 $\pm$ 0.029 & 0.909 $\pm$ 0.015 \\
&   & Logistic Regression & 0.670 $\pm$ 0.004 & 0.205 $\pm$ 0.006 & 0.899 $\pm$ 0.001 \\
\hline

\end{tabularx}
\caption{Cross-Validation scores of the global model on each site's local validation set. FedAvg and FedAvg\_HE achieved performance comparable to cML, whereas FedAvg\_DP showed performance degradation. Model architectures and hyperparameters were kept consistent across implementations to isolate privacy–utility trade-offs and avoid confounding from implementation-specific fine-tuning.}
\label{tab:global_model_privacy_methods}
\end{table}

\section*{Discussion}

Healthcare data is highly sensitive, and its exposure can lead to identity theft, discrimination, or other adverse consequences. FL reduces data exposure by keeping data local, but the sharing of model updates still poses privacy risks, as adversaries may infer sensitive information from gradients or parameters. To mitigate these risks, we integrated DP and HE into the FL workflow and evaluated their impact on model performance and system efficiency. We first simulated the FL implementations and then deployed them using a real-world setup. The three federated methods were FedAvg, FedAvg\_HE, and FedAvg\_DP. Unlike purely simulated federated settings, the system was deployed across independently managed environments, introducing real-world constraints such as network latency, resource heterogeneity, and decentralized system administration. Performance was benchmarked against a cML baseline for both NN and LR models. FedAvg and FedAvg\_HE matched cML in their discriminatory performance, demonstrating that FL can preserve model utility while enhancing privacy. In contrast, FedAvg\_DP exhibited reduced performance due to calibrated noise injection. Fine-tuning the SVTPrivacy algorithm further highlighted that low noise levels maintained performance close to cML, whereas higher noise levels caused instability and reduced accuracy.

Previous studies have primarily benchmarked DP using larger NN models~\cite{saifullah2024towards, wei2023dpmlbench, cheng2022dpnas, pannekoek2021investigating}. Our FedAvg\_DP implementation with the LR learner (only 11 trainable parameters) was more sensitive to noise than the NN learner (66 trainable parameters), demonstrating that DP hyperparameters must be tailored to the model scale and cannot be universally standardized. FedAvg\_HE preserved performance and achieved convergence similar to FedAvg, but introduced computational and communication overhead. Encrypted updates remained in kilobytes for LR but expanded to megabytes for NNs, suggesting that FedAvg\_HE is better suited to smaller models, while scalability becomes a concern for complex architectures. Since the encrypted updates remained in the megabyte range, the CKKS parameters were not further tuned; however, parameter optimization should be considered in future work as model complexity increases. Both methods present distinct trade-offs: FedAvg\_DP requires careful tuning as smaller models are particularly vulnerable to performance degradation from noise, whereas FedAvg\_HE maintains model fidelity but incurs overhead proportional to model size. In our clinical application, FedAvg\_DP-induced noise had a larger impact on predictive performance than the computational cost of HE, making FedAvg\_HE the more practical choice. Selection between the two should therefore account for model complexity, resource constraints, and privacy requirements. Although combining FedAvg\_HE and FedAvg\_DP could enhance privacy protection, it introduces substantial overhead and potential performance loss. Consequently, we evaluated FedAvg\_HE and FedAvg\_DP independently, leaving hybrid approaches for future studies to assess the security–efficiency trade-off.

This study focused on the technical implementation and evaluation of privacy-preserving FL methods. The deployed infrastructure was designed for technical validation only and was not integrated into clinical workflows or real-time healthcare environments. While the dataset was sufficient to demonstrate feasibility and trade-offs, it is not comprehensive for clinical risk prediction. Observed predictive performance aligned with general population baselines~\cite{CDT37250, hara2021claims, medicina61112042}, supporting the validity of our technical evaluation. However, limitations of our study include a restricted set of FL configurations, model architectures, aggregation strategies, and privacy parameters. Evaluation was confined to a single national dataset and horizontal FL, and therefore generalizability to other healthcare data, disease domains, or vertical/hybrid FL paradigms remains to be tested. Future research should explore adaptive privacy mechanisms, secure multi-party computation, trusted execution environments, synthetic data, and expanded evaluation metrics, including fairness, interpretability, and robustness.

In conclusion, this study demonstrated that privacy-preserving FL offers a viable pathway for collaborative model development in healthcare, enabling institutions to jointly train predictive models without compromising patient confidentiality. Through simulation and deployment, we integrated FL with differential privacy and homomorphic encryption, systematically comparing both approaches against cML in terms of predictive performance and computational efficiency. FedAvg\_HE emerged as the most suitable method for our use case, preserving model fidelity while providing strong privacy guarantees. The systematic comparison of FedAvg\_DP and FedAvg\_HE provided practical insights into the trade-offs between privacy protection, predictive performance, and computational overhead, which can guide implementation decisions in similar clinical settings. Overall, FL represents a scalable and practically deployable framework for sensitive domains such as healthcare, where the careful balance between data privacy and model utility remains a fundamental requirement.

\bibliography{sample}

\section*{Funding Statement}
This work was supported by the Business Finland-funded E! ITEA Secur-e-health project (4220/31/2021). 

\section*{Acknowledgements}
We acknowledge Success Clinic Oy for obtaining and providing Swedish data and Heba Sourkatti for her contributions to data processing, which facilitated the analyses in this study.

\section*{Author contributions statement}
G.S. designed the study, developed the methods, conducted experiments, analyzed the data, and drafted the initial manuscript. J.P. contributed to method development, implemented the SEH platform and conducted manuscript review. A.I. contributed to the methods and developed proof-of-concept implementations using NVFLARE. M.R. conducted the literature review and supported the development of homomorphic encryption methods. N.L. and P.A. hosted the remote clients and facilitated the real-world federated learning deployment. M.H. designed and supervised the study. All authors reviewed and revised the manuscript for intellectual content and approved the final version.

\section{Data and Code Availability}
The dataset was obtained from the National Board of Health and Welfare (Socialstyrelsen), a governmental agency under the Swedish Ministry of Health and Social Affairs responsible for maintaining national health registries.
The code is available at: \url{https://extgit.vtt.fi/gaurang.sharma/fl_with_pets}

\section{Ethics and Consent to Participate}
We used this anonymized patient data for statistical purposes; therefore, a formal data request to the National Board of Health and Welfare sufficed, and no additional ethical approval was required. All data processing adhered to the Health Data Register Act (1998:543) and its associated regulations and ordinances.

\section{Competing interests}
All authors declare no financial or non-financial competing interests.

\end{document}